\documentclass[times, review, 10pt]{elsarticle}
\usepackage{multirow}
\usepackage[letterpaper, top=4.3cm, bottom=4.3cm, left=4.8cm, right=4.8cm]{geometry}

\usepackage{mathptmx}  

\usepackage{graphicx}
\usepackage{amssymb}
\usepackage{amsmath}
\usepackage[linesnumbered,ruled,vlined]{algorithm2e}
\usepackage{booktabs}
\usepackage{makecell}
\usepackage[table]{xcolor}
\definecolor{lightorange}{HTML}{FFE6CC}
\usepackage{caption}
\renewcommand\footnotesize{\fontsize{8pt}{9.5pt}\selectfont}  

\usepackage[labelfont=bf, labelsep=period]{caption}

\usepackage{hyperref}
\usepackage{xcolor}
\usepackage[table]{xcolor}
\usepackage{setspace} 
\usepackage{float}
\journal{Pattern Recognition}

\begin{document}

\begin{frontmatter}

\title{\fontsize{14pt}{\baselineskip}\selectfont\selectfont Learn from Foundation Model:  Fruit Detection Model without Manual Annotation}


\author[label1]{Yanan Wang}
\author[label1,label2]{Zhenghao Fei}
\author[label3]{Ruichen Li}
\author[label1,label2]{Yibin Ying}

\affiliation[label1]{
    organization={College of Biosystems Engineering and Food Science, Zhejiang University},
    city={Hangzhou},
    postcode={310058},
    country={China}
}
\affiliation[label2]{
    organization={ZJU-Hangzhou Global Scientific and Technological Innovation Center, Zhejiang University},
    city={Hangzhou},
    postcode={311215},
    country={China}
}

\affiliation[label3]{
    organization={Faculty of Science, National University of Singapore},
    postcode={119077},
    country={Singapore}
}

\begin{abstract}
Deploying accurate and efficient detection models in complex scenes such as agriculture is critically hampered by the scarcity of large-scale, manually annotated datasets. To address this, we propose SDM-D, a prompt-driven framework that bypasses the need for manual instance-level annotation by distilling knowledge from large foundation models into lightweight, edge-deployable student models. Our approach starts with a novel, training-free SDM method to generate high-quality pseudo-labels, and is then integrated with a distillation mechanism to train compact student models. Comprehensive experiments demonstrate that our approach outperforms leading open-set methods like Grounded SAM2 and YOLO-World. Crucially, our zero-shot model reaches 86.6\% of the performance of its fully supervised counterpart, narrowing the gap to 91.6\% with just a single labeled example (one-shot fine-tuning). Furthermore, the distilled student models are also extremely efficient, achieving inference speeds over 100 times faster than the original FM pipeline, enabling real-time performance on edge devices. To facilitate evaluation and future research, we also introduce MegaFruits, a large, high-quality public dataset for fruit instance segmentation, comprising over 25,000 images. The code and dataset are publicly available at https://github.com/AgRoboticsResearch/SDM-D.git.
\end{abstract}

\begin{keyword}
Fruit Detection \sep Foundation Models \sep Knowledge Distillation \sep Zero-Shot Learning \sep Agriculture
\end{keyword}
\end{frontmatter}

\begin{figure*}[!t]
    \centering
    \includegraphics[width=1\textwidth]{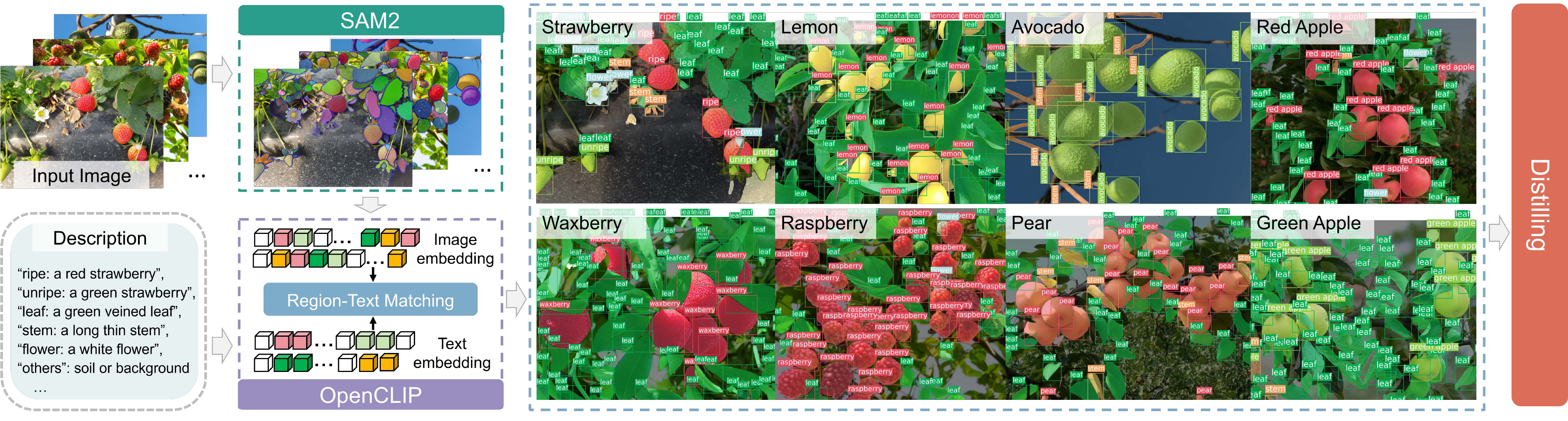}
    \small
    \caption[short caption]{SDM-D can simultaneously detect and segment input images based on the prompts, and enable distillation of knowledge from foundation models to faster, smaller models.}
    \label{fig1:SDM-D}
    \vspace{-0.3cm}
\end{figure*}



\section{Introduction}
Fruit production is vital for ensuring food security, feeding a growing global population, and many of the UN’s Sustainable Development Goals\cite{1_USDA_agricultural}. In 2022, fruits accounted for 10\% of global primary crop output, with nearly one billion tonnes produced\cite{2-fao_yearbook_2024}. Fruit visual perception technologies have been proven to be integral across various stages of fruit production\cite{pr-tomato3}, enabling precise yield estimation, disease identification to avoid damage, providing the necessary information for robotic picking, and more, which is not merely an economic or technological endeavor but a vital step towards securing the future of humanity.

In the last few years, deep learning (DL) methods have emerged as a mainstream approach in agricultural visual perception \cite{3-Vougioukas2019AgriculturalR}, however, training high-performance networks frequently requires vast datasets, often comprising thousands of images, and even state-of-the-art (SOTA) DL detection methods falter under limited data \cite{pr-DL}. The collection of large-scale, general-purpose datasets has greatly accelerated research and development in fields such as computer vision and autonomous driving\cite{10-1405.0312}. However, in the domain of agriculture, factors including environmental variability, crop seasonality, and concerns about data ownership pose challenges to the creation of large, diverse datasets. 
Existing fruit datasets (summarized in Table \ref{table:fruit_datasets}) are often unsuitable for the demands of modern OVS research. Many provide only bounding boxes for detection, while those with segmentation masks are typically small in scale (e.g., DeepBlueberry \cite{29-8787818}) or adopt highly specialized annotation criteria (e.g., MinneApple\cite{18-hani_minneapple_2020}), limiting their utility for training generalizable models. This data scarcity makes it exceedingly difficult to develop robust fruit perception systems that can operate across different crops and environments without laborious, per-instance manual annotation for each new domain. Therefore, developing an efficient fruit detection model generation method without the need for manual instance-level annotation is an open problem at present.

Recent breakthroughs in foundation models (FMs), such as the Segment Anything Model (SAM) \cite{21-kirillov_segment_2023} and CLIP\cite{23-radford2021learningtransferablevisualmodels}, have revolutionized computer vision by demonstrating powerful zero-shot generalization capabilities derived from training on massive datasets. After training on 11 million images with over 1 billion masks, SAM demonstrates zero-shot segmentation that generalizes exceptionally well to unfamiliar objects and images. However, the resources required to produce these models are immense, creating a significant barrier to their direct application.
For instance, GPT-3, a 175-billion-parameter “outdated” large language model, is estimated to have cost approximately \$4.6 million to train and would take 355 years on a single GPU.\cite{26-noauthor}. Similarly, CLIP's training relied on a dataset of 400 million image-text pairs, with its largest model requiring 18 days on 592 V100 GPUs\cite{23-radford2021learningtransferablevisualmodels}. This prohibitive cost for both training and deployment limits their use in specialized, resource-constrained domains. This challenge is particularly acute in agriculture, which presents two problems: data scarcity and the need for real-time, on-device processing, together with inherently complex visual conditions. Agricultural robots must perceive dense scenes (e.g., fruit clusters), where heavy occlusion and instance overlap are common. Under such cluttered conditions, prompt-driven open-vocabulary segmentation (OVS) paradigms often break down, resulting in missed or redundant detections (see Fig. \ref{fig:seg-then-pro} and Section 3 for the detailed comparison). This study addresses a critical question: How can the powerful, generalized knowledge of large FMs be effectively transferred to a lightweight, efficient model suitable for edge deployment in agriculture, without requiring extensive manual instance-level annotation? To answer this, we introduce SDM-D, a framework designed to distill knowledge from FMs for the comprehensive segmentation of complex agricultural scenes. The key contributions of this paper are as follows:
\begin{itemize}
    \item We propose SDM, a novel training-free, prompt-driven segment-then-prompt paradigm that leverages the pre-trained knowledge of foundation models to perform robust zero-shot segmentation in complex scenes.
    \item We present an effective knowledge distillation pipeline that enables the generation of compact student models for downstream tasks without requiring manual instance-level annotation. This approach facilitates the deployment of foundation model knowledge on edge devices for real-world field applications.
    \item  We provide comprehensive experimental evaluations demonstrating that our distilled models significantly outperform leading open-vocabulary baselines and offer substantial annotation savings, delivering a zero-shot performance comparable to that of models requiring 200 manually annotated images for training.
    \item To address data scarcity and facilitate future research, we contribute and open-source the MegaFruits dataset. With over 25,000 images, it is the largest public fruit instance segmentation dataset to our knowledge, and is released alongside all our code.
\end{itemize}

\begin{table*}[!t]
\centering
\caption{\footnotesize List of publicly available fruit detection datasets and our MegaFruit dataset.}
\label{table:fruit_datasets}
\resizebox{0.98\textwidth}{!}{%
\begin{tabular}{llllll}
\toprule
\textbf{Datasets} & \textbf{Annotation categories} & \textbf{Images} & \textbf{Instances} & \textbf{Labels} & \textbf{Task} \\
\midrule
MangoYOLO \cite{28-article} & Mango & 1,730 & 9,067 & Bounding-box & Object detection \\
DeepBlueberry \cite{29-8787818} & Blueberry & 7 & 228 & Mask & Instance segmentation \\
 &  & 293 & 10,161 & Bounding-box & Object detection \\
KFuji RGB-DS \cite{30-PMID:31406905} & Apple & 967 & 12,839 & Bounding-box & Object detection \\
MetaFruit \cite{31-li_metafruit_2024} & Apple, Orange, Lemon, Tangerine, Grapefruit & 4,248 & 248,015 & Bounding-box & Object detection \\
MinneApple \cite{18-hani_minneapple_2020} & Apple & 1,001 & 41,325 & Mask & Instance segmentation \\
StrawDI\_Db1 \cite{17-perez-borrero_fast_2020} & Ripe-Straw, Unripe-Straw & 3,100 & 17,938 & Mask & Instance segmentation \\
MegaFruits (ours)  & Starwberry, Blueberry, Peach & 25,182 & 87,952 & Mask & Instance segmentation \\
\bottomrule
\end{tabular}
}
\end{table*}

\section{Related Work}
\subsection{Open-vocabulary Detection}
    Traditional fruit detection methods primarily involve training closed-set detection models on specifically collected and labeled datasets. Consequently, these models can only respond to objects within a fixed set of labeled categories. Their performance is constrained by the size and quality of manually annotated datasets, and they typically cannot generalize well to unfamiliar domains. The rise of FMs has shifted the focus toward open-vocabulary object detection (OVD), which can detect objects in categories not explicitly labeled during training and generalize to unfamiliar images \cite{33-10457133}. This shift is particularly beneficial in the field of agriculture, where labeled data are often scarce. Grounding DINO \cite{36-liu_grounding_2024} advanced OVD by incorporating referring expression comprehension (REC) \cite{37-wu_referring_2022}, which is crucial for scenarios where objects are described based on their properties. This advancement aids in distinguishing between objects of the same category. While these models excel in OVD, they can not handle pixel-level segmentation tasks. Grounded SAM \cite{40-ren_grounded_2024} presented an innovative combination of the open-set detector Grounding DINO \cite{36-liu_grounding_2024} with the foundation segmentation model SAM \cite{21-kirillov_segment_2023}. This approach effectively addresses open-set segmentation tasks by initially conducting object detection based on the input text prompt, and then performing segmentation using the detection outputs. Similarly, YOLO-World \cite{41-cheng_yolo-world_2024} employs CLIP \cite{23-radford2021learningtransferablevisualmodels} for text encoding within a YOLO structure \cite{42-10533619}, achieving high inference speeds for open-set detection.

\subsection{Application of FMs in Agriculture}
There are some initial studies focused on deploying FMs in agriculture. This study \cite{43-yang_sam_2023} evaluated SAM's zero-shot segmentation on chickens using part-based and infrared thermal images, finding it outperformed SegFormer \cite{44-xie_segformer_2021} in both whole and part-based chicken segmentation. This work \cite{31-li_metafruit_2024} fine-tuned Grounding DINO on MetaFruit for open fruit object detection, demonstrating its impressive adaptability in learning. Nevertheless, this system lacks the capability for fruit segmentation and the performance of inference speed is still limited. To the best of our knowledge, in fruit segmentation, an effective framework for training a well-performed model without manual instance-level annotation is still lacking.

\subsection{Knowledge Distillation}

Another related field to this study is knowledge distillation. Although FMs generalize well to unfamiliar domains and tasks, they often need substantial computational resources, making them challenging to deploy efficiently on edge devices such as robots. Knowledge distillation has been explored to address these issues \cite{48-kozlov_neural_2021}. In knowledge distillation, a "teacher" model transfers its knowledge to a smaller "student" model, enabling the student to achieve comparable performance while being more resource-efficient \cite{50-Hinton2015DistillingTK}. In a typical knowledge distillation process, the student model being trained to mimic the output probabilities (or logits) of the teacher model, and a loss function is used to measure the gap between the student's and teacher's predictions. Additionally, Xie et al. \cite{51-xie2020selftrainingnoisystudentimproves} demonstrated that distillation could be achieved by propagating pseudo-labels to unlabeled data in a self-supervised pipeline, linking knowledge distillation to pseudo-labeling without relying on output matching. This establishes an important connection between knowledge distillation and pseudo-labeling. Our work builds on this relation and extends knowledge distillation to scenarios where no manual labels are available.

\section{Methodology}
\begin{figure}
    \centering
    \includegraphics[width=0.9\textwidth]{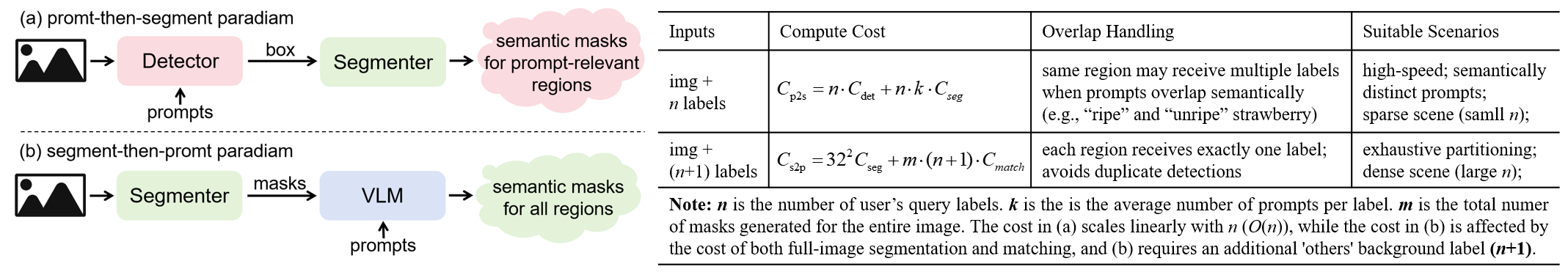}
    \small
    \caption[short caption]{Comparison of two paradigms for open-vocabulary segmentation. (a) Prompt-then-segment: images and prompts are jointly processed by a detector to produce prompt-related regions, which are refined by a segmenter into prompt-specific masks. (b) Segment-then-prompt: the image is first segmented into category-agnostic masks, which are then matched with prompts to assign semantics to all regions.}
    \label{fig:seg-then-pro}
    \vspace{-0.3cm}
\end{figure}

To address challenges in dense agricultural scenes, we propose a novel framework. Conventional open-vocabulary models like Grounded SAM Grounded SAM \cite{40-ren_grounded_2024} and YOLO-World \cite{41-cheng_yolo-world_2024} employ a “prompt-then-segment” paradigm, as illustrated in Fig. \ref{fig:seg-then-pro}(a). In this workflow, text prompts first guide a detector to identify relevant object regions, which are subsequently passed to a segmenter. While this approach is efficient, the final segmentation in this paradiam is conditioned on the initial prompt-based detections, which is prone to errors of omission (missing valid objects not covered by prompts) and commission (generating duplicate masks for a single instance). To overcome, our SDM framework introduces a “segment-then-prompt” paradigm (Fig. \ref{fig:seg-then-pro}(b)). It reverses the workflow by first generating a comprehensive set of category-agnostic masks for all potential objects and then matching them with text prompts. This decoupling of segmentation from detection effectively mitigates the issues of missed and duplicate detection. Furthermore, we integrate SDM into a knowledge distillation pipeline, termed SDM-D  (Fig. \ref{fig:2-flow-SDM}),  to obtain lightweight student models suitable for real-time inference on edge devices.

\subsection{Segmentation}

To segment all targets as accurately as possible, we use SAM2 as the dominant algorithm, which is a prompt-driven segmentor that comprises an image encoder, a prompt encoder, and a lightweight mask decoder. The image encoder is a masked autoencoder pre-trained ViT64 minimally adapted to process high-resolution inputs. Input an image with a resolution, the image encoder will process it as Equation \ref{eq1}:

\begin{equation}
\label{eq1}
E_I = \epsilon_{\text{img}}(I) \in \mathbb{R}^{\left( \frac{H}{16} \times \frac{W}{16} \right) \times d}
\end{equation}
where $E_I$ is the image embedding of $I$, $d$ is the feature dimension. The image encoder runs once per image and can capture high-level semantic information while maintaining spatial relationships within the image. Given the complexity of agricultural scenes and the need for method generalization, we opt to use a set of $32 \times 32$ regular grid points $P = \{p_1, p_2, \ldots, p_{32}\}$ as the input prompts, where each point $p_i = (x_i, y_i)$ specifies a location in the image. Then these points are embedded using positional encodings $\pi(p_i)$ and point-specific learned embeddings $v_p$ to get point embeddings $E_P$, as Equation (\ref{eq:2}):

\begin{subequations}\label{eq:2}
\begin{align}
e_i &= \pi_j\bigl(p_i\bigr) + v_p \, e_i \;\in\; \mathbb{R}^d \label{eq:2A}\\
E_P &= \{\,e_i \mid i = 1, 2, \ldots, m\}, 
\quad E_p \;\in\; P^{m \times d} \label{eq:2B}
\end{align}
\end{subequations}
Each point in the grid is mapped to a 256-dimensional vectorial embedding. Rather than relying on task-specific or manually defined prompts, these point grids can simplify the process and enhance the scalability of the model across different image types, what’s more, we emphasize that the user can customize the number of grid points according to the density of the objects in the image. The two-layer mask decoder, using prompt self-attention and cross-attention, then maps the image embedding $E_I$ and prompt embeddings $E_p$ to a set of masks $M_i$, as Equation (\ref{eq4}):

\begin{equation}
\label{eq4}
M = f_{\text{mask}}(E_I, E_P)
\end{equation}
where $M = \{ M_1, M_2, ..., M_n \}$, and each $M_i$ is defined as the intersection of the original image and a binary mask. 
Since SAM2 is a model with ambiguity-awareness, it allows the model to generate multiple candidate masks when it encounters uncertain or overlapping regions and then returns the top three mask outputs based on loss ranking. To avoid generating small, irrelevant regions, we eliminate all unconnected regions with less than 50 pixels, which are determined by the characteristics of our agriculture datasets. However, when a grid lies on a part or sub-part of objects, multiple masks may still be generated. In such cases, we keep all the segmentation results and remove the ambiguity in post-processing.

\subsection{Mask NMS}
The final output of SAM2 often results in multiple masks with significant overlap or redundancy, where an object can appear in multiple masks, or a single mask may cover multiple objects. These issues are particularly common in fruit images, such as strawberries with calyxes (see Fig.~2(a)). They can significantly affect the downstream tasks in robotic operations, leading to incorrect grasping, damage to fruit, or failure to identify the fruit entirely, which in turn lowers operational efficiency and increases waste. To address these challenges, we propose a \textit{Mask NMS} (Non-Maximum Suppression) mechanism, which is specifically designed to retain the optimal mask that covers only a single fruit instance (e.g., mask2 in Fig.~2(a)) and eliminate ambiguity.
The decision criterion is shown in Equation (\ref{eq5}). Unlike traditional NMS, which relies on bounding box IoU, our approach calculates the overlap area between each pair of masks. If the overlap ratio, based on the smaller mask, surpasses the pre-defined confidence threshold, we retain the mask with the higher score.

\begin{equation}
\text{$M$} = 
\begin{cases}
M_i, & \text{if } \frac{|M_i \cap M_j|}{|M_i|} > C \text{ and } S(M_i) > S(M_j) \\
M_j, & \text{else}
\end{cases}
\label{eq5}
\end{equation}
where $M_i$ presents the mask with a smaller area. $C$ is the predefined confidence threshold, which was set at $0.9$ in our experiment. $S_i$ is the stability\_score of the $i^{\text{th}}$ mask output by SAM2.

\subsection{Description and Image Segments Encoding}

To facilitate seamless image–text comparisons, we employ OpenCLIP\cite{24-ilharco_openclip_2021}. And to achieve more accurate object recognition and classification—particularly in agricultural scenarios where objects often exhibit subtle visual differences and may be affected by environmental factors (e.g., a fully red ripe strawberry vs. a half-red unripe strawberry)—we incorporate region-level descriptions following REC\cite{37-wu_referring_2022}. As illustrated in Fig. \ref{fig:2-flow-SDM}(c), instead of inputting only the category labels of the instances, the user provides descriptive prompts for the categories of interest (e.g., “a red strawberry” instead of “a strawberry”). These structured descriptions generate feature-rich embeddings that improve the alignment between visual features and textual semantics. Since SDM assigns a label to every segmented region, we additionally introduce an “others” category to handle irrelevant regions or background areas. This ensures comprehensive labeling while maintaining focus on the target categories. The design of prompt templates and their impact on performance are further discussed in Section 4.6.2. OpenCLIP’s dual-encoder setup, coupled with its extensive pre-training on large image-text pairs, enables it to effectively handle a large vocabulary, including out-of-vocabulary words. For example, given an object category, its corresponding text description $t_c$ is processed through the text encoder to produce a text embedding $t_c$ as Equation \eqref{eq:6}:
\begin{equation}
    t_c = f_{\text{text}}(t)
    \label{eq:6}
\end{equation}
where $f_text$ is the text encoder of Open CLIP. This process enables the model to recognize and classify objects that may not have been seen during training, significantly improving performance in real-world applications.
Simultaneously, each predicted segment is processed to generate a corresponding image embedding. For every segmentation mask $m_i \in M$, we use it to crop the object region from the original image $I$. This operation effectively isolates the object by removing all background pixels. These background-free, cropped images are then individually fed into the OpenCLIP image encoder, and the image embeddings are computed as shown in Equation \eqref{eq:image_embedding}:

\begin{equation}
\label{eq:image_embedding}
E_M = f_{\mathrm{img}}(I \odot M_i), \quad i = 1, 2, \dots, k
\end{equation}
where $E_M = \{e_1, e_2, \cdots, e_k\}$ contains $k$ image embeddings, $f_{\text{img}}$ represents the image encoder of OpenCLIP that maps the masked image to the embedding space, and $\odot$ is element-wise multiplication, restricting $I$ to the masked region defined by $M_k$. This mask-based cropping strategy is crucial. It forces the encoder to focus solely on the features of the objects themselves, thereby reducing computational complexity and minimizing interference from background clutter (see Appendix A.3 for more details).

\subsection{Region-Text Matching}
\begin{figure*}[ht]
    \centering
    \includegraphics[width=0.85\textwidth]{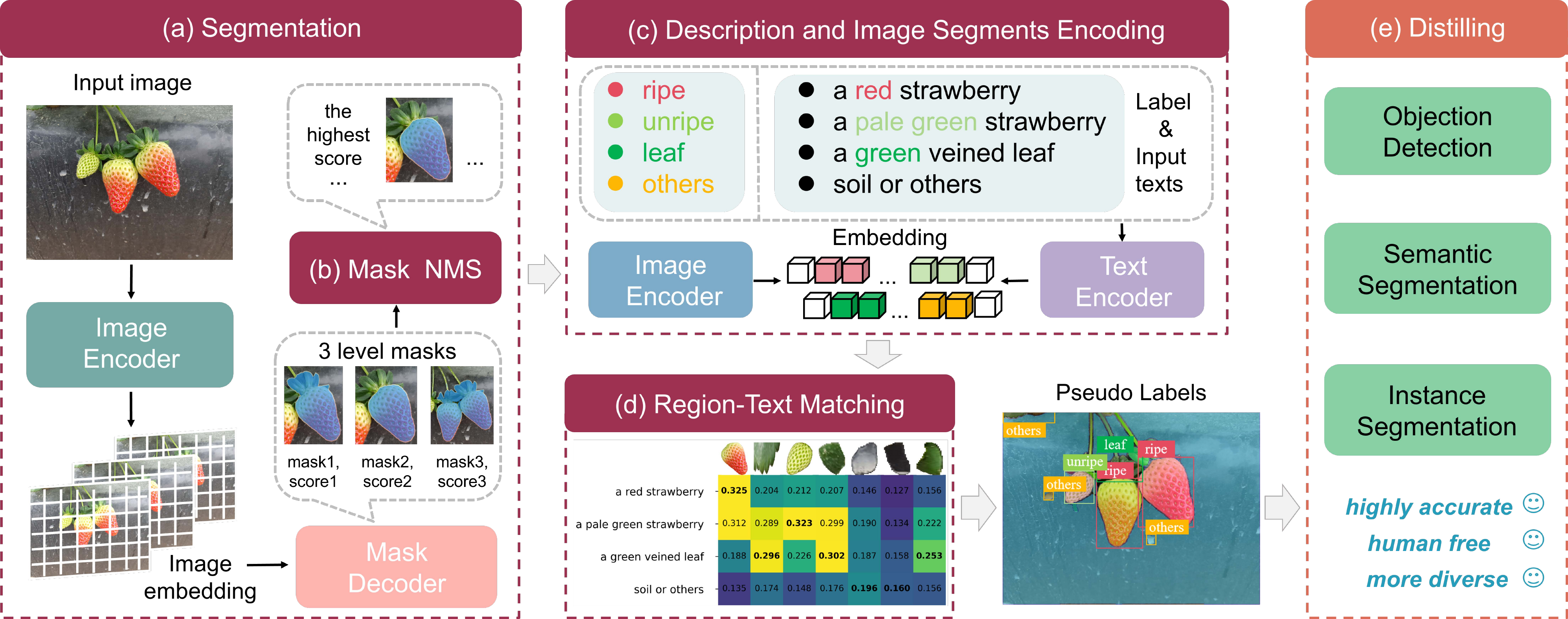}
    \small
    \caption[short caption]{Overall framework of SDM-D. (a) Segmentation: A SAM2-based image encoder-decoder generates dense region proposals. (b) Mask NMS: was proposed to reserve the optimal mask. (c) Description and Image Segments Encoding: OpenCLIP is used to encode the description and image segments. (d) Region-Text Matching: OpenCLIP will calculate the similarity between the image regions and textual description to select the matching. (e) Distilling: involves transferring knowledge to smaller models that are faster and perform better.}
    \label{fig:2-flow-SDM}
    \vspace{-0.3cm}
\end{figure*}

To align image regions with textual descriptions, we adopt OpenCLIP \cite{24-ilharco_openclip_2021}, which establishes a powerful multimodal embedding space through large-scale pre-training on 400 million image-text pairs. Given a batch of $N$ image-text pairs, the model considers all $N \times N$ possible pairings between image and text embeddings. The training objective of OpenCLIP is to maximize the cosine similarity of the $N$ true pairs while simultaneously minimizing the similarity of the remaining $N^2 - N$ mismatched pairs. In this setting, cosine similarity is not a trivial distance metric but the very foundation of CLIP's contrastive learning objective, ensuring that semantically aligned image-text pairs are mapped close together in the embedding space. Specifically, given a set of text embeddings $T$=\{$t_1$, $t_2$, \dots, $t_c$\} for $C$ textual descriptions, and a region embedding $e$, the classification probabilities between them are computed as Equation \eqref{eq:7}:

\begin{equation}
    P = \frac{T \cdot e}{\|e\|}
    \label{eq:7}
\end{equation}
where $P$=\{$p_1$, $p_2$, \dots, $p_c$\} inner product of the vector, $\cdot$ is the dot product, and $\| \|$ is the calculation of the Euclidean norm. The label of each $r$ is then returned by the index of maximum similarity, represented as equation (\ref{eq8}):

\begin{subequations}\label{eq8}
\begin{align}
i &= \arg\max_{c \in \{1,2,\dots,C\}} \langle x, L_c \rangle p_c \label{eq:2A} \\
y &= L_i \label{eq:2B}
\end{align}
\end{subequations}
where $i$ is the index of the maximum similarity score in $P$, and $L$ = \{$l_1$, $l_2$, \dots, $l_1$\} is a label set corresponding to $P$. By jointly training an image encoder and text encoder to learn a multi-modal embedding space, OpenCLIP can maximize the cosine similarity of the image and text embeddings of the $N$ real pairs in the batch while minimizing the cosine similarity of the embeddings of the $N^2$ - $N$ incorrect pairs. As shown in Fig.\ref{fig:2-flow-SDM}(d), we also inherit the matrix representation from CLIP, providing an intuitive interpretation of the match. This method enhances the understanding of image-text relationships, supporting advanced applications like image captioning, visual question answering, and semantic segmentation in complex agricultural environments.

\subsection{Distilling}
To facilitate efficient deployment on edge devices, we implement a practically novel distillation process. Given a dataset $D$ = \{$i_1$, $i_2$, \dots, $i_n$\}, SDM is used to generate pseudo-labels for each image, represented as Equation \eqref{eq:11}:
\begin{equation}
    Y_p = f(D, T)
    \label{eq:11}
\end{equation}
where $ Y_p = \{y_p^1, y_p^2, \dots, y_p^n\} $ represents the set of $ n $ pseudo-labels predicted by SDM in a zero-shot manner. $ D $ denotes the dataset containing images, and $ T $ represents the set of textual descriptions (prompts). These pseudo-labels serve as supervision for small edge-deployable models (students), bypassing the need for costly manual annotation. Unlike traditional distillation which typically operates at the feature or logit level using manually labeled data, our approach performs distillation at the label level via pseudo-labels. This strategy treats the final pseudo-labels generated by the SDM pipeline as the ground truth for the student. Consequently, any cumulative errors from the upstream foundation models (e.g., SAM, OpenCLIP) are inherently embedded within this supervisory signal. This process can be viewed as a form of self-supervised learning, where the student model's parameters, $\theta^*$, are optimized by minimizing a total loss function, $L_{\text{total}}$, between its predictions and the SDM's pseudo-labels. The process can be expressed as Equation \eqref{eq:12}:

\begin{subequations}\label{eq:12}
\begin{align}
\theta^*&= \arg\min_{\theta} \sum_{k=1}^{t}  \label{eq:2A}\\
\hat{y}_p^k &=f_{\theta}(i_k) \label{eq:2B}
\end{align}
\end{subequations}
where $ {y}_p^k $ represents the $ k^{th} $ label predicted by SDM, and $ \hat{y}_p^k $ is the corresponding prediction from the student model.
The composition of $L_{\text{total}}$ is determined by the architecture of the chosen student model, making our framework flexible. To provide a concrete example, we define $L_{\text{total}}$ for the instance segmentation task using a YOLOv8 student model. In this case, the loss is the standard weighted sum of its native components. The overall loss is formulated as Equation \eqref{eq:yolov8_loss_components}:

\begin{subequations}
\label{eq:yolov8_loss_components}
\begin{align}
    L_{\mathrm{cls}} &= -\frac{1}{N} \sum_{i=1}^{N} [y_i \log \hat{p}_i + (1 - y_i) \log(1 - \hat{p}_i)] \label{eq:loss_cls} \\
    L_{\mathrm{box}} &= 1 - \mathrm{IoU}(b, \hat{b}) + \frac{\rho^2(b_{\mathrm{center}}, \hat{b}_{\mathrm{center}})}{c^2} + \alpha v \label{eq:loss_box} \\
    L_{\mathrm{dfl}} &= -\frac{1}{N} \sum_{i=1}^{N} \sum_{j=1}^{K} y_{i,j} \log \hat{p}_{i,j} \label{eq:loss_dfl} \\
    L_{\mathrm{mask}} &= -\frac{1}{M} \sum_{u=1}^{M} [m_u \log \hat{m}_u + (1 - m_u) \log(1 - \hat{m}_u)] \label{eq:loss_mask} \\
    L_{\text{total}} &= \lambda_{cls}L_{cls} + \lambda_{box}L_{box} + \lambda_{dfl}L_{dfl} + \lambda_{mask}L_{mask} \label{eq:loss_total}
\end{align}
\end{subequations}
where $L_{\mathrm{cls}}$, $L_{\mathrm{box}}$, and $L_{\mathrm{dlf}}$ are the classification, bounding box (CIoU), and distribution focal loss terms native to the YOLOv8 detection head. $L_{\mathrm{mask}}$ is the pixel-wise binary cross-entropy loss for the segmentation mask head. The scalar balancing coefficients $\lambda_{cls}$, $\lambda_{box}$, $\lambda_{dfl}$, and $\lambda_{mask}$ are adopted from the standard YOLOv8-Seg implementation. Training halts based on the early stopping mechanism if the validation loss does not decrease for 100 epochs. 
Notably, student models consistently outperform SDM models on specific tasks, ensuring distilled model predictions better approximate ground truth values - critical for real-time agricultural applications requiring high accuracy and low computational demands. Furthermore, the method is model-agnostic, allowing any compact model optimized for downstream tasks to seamlessly integrate into the distillation process. This approach bridges the gap between powerful foundation models and practical small models, enabling efficient knowledge transfer.

\section{Experiments and Results}
\subsection{Datasets and Metrics}

\subsubsection{MegaFruits Dataset}
To address the scarcity of high-quality public datasets for open-vocabulary segmentation (OVS) in agriculture, we introduce MegaFruits: a large-scale, richly annotated dataset designed for fruit segmentation tasks. A comprehensive overview of all datasets used in our experiments, detailing their roles and annotation status, is provided in Table \ref{tab:dataset_distribution}.

\begin{table}[h]
\centering
\footnotesize
\caption{The distribution table for datasets used in the experiment.}
\label{tab:dataset_distribution} 
\resizebox{\textwidth}{!}{
{
\begin{tabular}{lcccccccccccccc} 
\toprule
\ & \multicolumn{4}{c}{\textbf{StrawDI\_Db1}} & \multicolumn{4}{c}{\textbf{Mega\_Blueberry}} & \multicolumn{4}{c}{\textbf{Mega\_Peach}} & \multicolumn{2}{c}{\textbf{MegaStrawberry}} \\ 
\cmidrule(lr){2-5} \cmidrule(lr){6-9} \cmidrule(lr){10-13} \cmidrule(lr){14-15} 
& Train & Val & Test & Total & Train & Val & Test & Total & Train & Val & Test & Total & Test & Train* \\ 
\midrule
Images & 2,800 & 100 & 200 & 3,100 & 1,778 & 254 & 508 & 2,540 & 1,680 & 240 & 480 & 2,400 & 290 & 19,952 \\ 
Instances & 16,234 & 572 & 1,132 & 17,938 & 12,898 & 1,802 & 5,956 & 20,656 & 6,349 & 980 & 2,800 & 10,129 & 5,695 & 518,712 \\ 
\bottomrule
\end{tabular}
}}
\vspace{0.5em}
\parbox{0.8\textwidth}{
\footnotesize
\textbf{Note:} The train set of MegaStrawberry* is pseudo labels from SDM.}

\end{table}

The MegaFruits dataset comprises three distinct subsets: MegaBlueberry, MegaPeach, and MegaStrawberry. Data collection was conducted between October 20, 2023, and July 6, 2024, in real-world agricultural environments without artificial modifications, ensuring that the data reflects authentic challenges. All test data were independently sourced from distinct orchard rows or regions to mitigate location-specific biases and ensure no overlap with the data used for training and validation. What’s more, the MegaFruits naturally captures a wide array of challenging conditions, including dense fruit clusters, varying illumination (e.g., direct sunlight, shadows, backlight), occlusion from leaves and stems, and fruit adhesion. In particular, the MegaStrawberry subset was sourced from three different orchards, encompassing both elevated and ground-based cultivation methods. Example images from MegaStrawberry are displayed in Fig. \ref{fig:MegaFruits}. The annotation protocol for MegaFruits is designed to support both large-scale model training and rigorous, unbiased evaluation. The entirety of the MegaBlueberry and MegaPeach subsets was manually annotated by trained experts using precise polygonal masks. For the MegaStrawberry subset, we adopted a hybrid strategy: a large-scale training set was generated using high-quality pseudo-labels from our SDM pipeline, while a dedicated test set of 290 images was manually annotated to serve as the ground truth. This approach allows for the creation of extensive training data at low cost while ensuring that model evaluation is performed against a reliable, human-verified standard. In addition to MegaFruits, we also utilize the public StrawDI\_Db1 dataset \cite{17-perez-borrero_fast_2020}. For consistency in our experiments, we merge its “rip” and “unrip” strawberry classes into a single “strawberry” category.

\begin{figure}
    \centering
    \includegraphics[width=0.8\textwidth]{dataset1-esp-converted-to.pdf}
    \small
    \caption[short caption]{Representative examples of MegaFruits: (a) Object detection. (b) Semantic segmentation. (c) Instance segmentation task. (d) MegaStrawberry train set with pseudo masks generated by SDM.}
    \label{fig:MegaFruits}
    \vspace{-0.3cm}
\end{figure}

\subsubsection{Evaluation Metrics}
For object detection and instance segmentation, we adhere to the standard evaluation metrics established by COCO \cite{10-1405.0312}, focusing on three key metrics: mAP$_{50:95}$, mAP$_{50}$, and mAR. For semantic segmentation, we employ the VOC \cite{56-everingham_pascal_2010} evaluation metrics, focusing on class accuracy, mIOU, and FWIOU.

\subsection{Zero-shot Open-vocabulary Perception}

\begin{figure}
    \centering
    \includegraphics[width=0.5\textwidth]{Figure_4-eps-converted-to.pdf}
    \small
    \caption[short caption]{Comparison of zero-shot performance. (a) Comparison of fruit object detection results on a strawberry image. (b) Comparison of fruit semantic segmentation results on a blueberry image. (c) Comparison of fruit instance segmentation results on a peach image.}
    \label{fig:zero-det}
    \vspace{-0.3cm}
\end{figure}

\subsubsection{Quantitative Performance Evaluation}
In this section, we evaluated the zero-shot open-vocabulary performance of our training-free SDM method against three state-of-the-art (SOTA) baselines. We compared the direct predictive outputs of these methods against manual annotations across three tasks: object detection, semantic segmentation, and instance segmentation. The quantitative results are presented in Table \ref{tab2:zeroshot_results}, with a visual comparison in Fig. \ref{fig:zero-det}.

As illustrated in Table \ref{tab2:zeroshot_results}, SDM demonstrates superior performance by a large margin across all tasks. For instance segmentation, SDM  achieves 1.85, 1.59, and 1.79 times the mAP$_{50:95}$ performance of the second-best method on the strawberry, blueberry, and peach datasets, respectively. This substantial improvement underscores SDM’s capability complex general fruit scenes at the level of individual fruits and to capture fine-grained boundary details.

\begin{table*}[!t]
\centering
\footnotesize
\caption{The zero-shot performance of SDM on three tasks.}
\label{tab2:zeroshot_results}
\resizebox{0.8\textwidth}{!}{%
\begin{tabular}{l ccc ccc ccc}
\toprule
\multicolumn{10}{c}{\textbf{Results of zero-shot object detection}} \\
\midrule
\textbf{Method}
 & \multicolumn{3}{c}{\textbf{StrawDI\_Db1}} & \multicolumn{3}{c}{\textbf{Mega\_Blueberry}} & \multicolumn{3}{c}{\textbf{Mega\_Peach}} \\
\cmidrule(lr){2-4} \cmidrule(lr){5-7} \cmidrule(lr){8-10}
 & mAP$_{50:95}$ & mAP$_{50}$ & mAR & mAP$_{50:95}$ & mAP$_{50}$ & mAR & mAP$_{50:95}$ & mAP$_{50}$ & mAR \\
\midrule
Grounded SAM & 0.216 & 0.322 & 0.401 & 0.232 & 0.251 & 0.356 & 0.190 & 0.228 & 0.220 \\
Grounded SAM2 & \underline{0.290} & \underline{0.347} & \underline{0.497} & \underline{0.251} & 0.282 & 0.389 & 0.271 & 0.244 & \underline{0.473} \\
YOLO-World & 0.173 & 0.232 & 0.429 & 0.233 & \underline{0.314} & \underline{0.517} & \underline{0.287} & \underline{0.410} & 0.335 \\
SDM (Ours) & \textbf{0.540} & \textbf{0.635} & \textbf{0.639} & \textbf{0.411} & \textbf{0.462} & \textbf{0.633} & \textbf{0.524} & \textbf{0.596} & \textbf{0.686} \\
\midrule
\multicolumn{10}{c}{\textbf{Results of zero-shot semantic segmentation}} \\
\midrule
\textbf{Method}
 & \multicolumn{3}{c}{\textbf{StrawDI\_Db1}} & \multicolumn{3}{c}{\textbf{Mega\_Blueberry}} & \multicolumn{3}{c}{\textbf{Mega\_Peach}} \\
\cmidrule(lr){2-4} \cmidrule(lr){5-7} \cmidrule(lr){8-10}
 & Class Acc. & mIOU & FWIOU & Class Acc. & mIOU & FWIOU & Class Acc. & mIOU & FWIOU \\
\midrule
Grounded-SAM & 0.936 & 0.832 & 0.959 & 0.614 & 0.553 & 0.829 & 0.683 & 0.644 & 0.858 \\
Grounded SAM2 & \underline{0.941} & \underline{0.857} & \underline{0.969} & \underline{0.628} & \underline{0.571} & \underline{0.851} & 0.836 & 0.801 & 0.892 \\
YOLO-World & 0.863 & 0.768 & 0.944 & 0.621 & 0.536 & 0.820 & \underline{0.901} & \underline{0.875} & \underline{0.948} \\
SDM (Ours) & \textbf{0.959} & \textbf{0.917} & \textbf{0.981} & \textbf{0.813} & \textbf{0.760} & \textbf{0.901} & \textbf{0.914} & \textbf{0.882} & \textbf{0.951} \\
\midrule
\multicolumn{10}{c}{\textbf{Results of zero-shot instance segmentation}} \\
\midrule
\textbf{Method}
 & \multicolumn{3}{c}{\textbf{StrawDI\_Db1}} & \multicolumn{3}{c}{\textbf{Mega\_Blueberry}} & \multicolumn{3}{c}{\textbf{Mega\_Peach}} \\
\cmidrule(lr){2-4} \cmidrule(lr){5-7} \cmidrule(lr){8-10}
 & mAP$_{50:95}$ & mAP$_{50}$ & mAR & mAP$_{50:95}$ & mAP$_{50}$ & mAR & mAP$_{50:95}$ & mAP$_{50}$ & mAR \\
\midrule
Grounded SAM & 0.247 & \underline{0.376} & 0.438 & 0.234 & 0.255 & 0.355 & 0.175 & 0.212 & 0.212 \\
Grounded SAM2 & \underline{0.297} & 0.362 & \underline{0.504} & 0.256 & 0.288 & 0.385 & \underline{0.253} & \underline{0.316} & 0.443 \\
YOLO-World & 0.119 & 0.222 & 0.239 & \underline{0.258} & \underline{0.319} & \underline{0.511} & 0.163 & 0.234 & \underline{0.622} \\
SDM (Ours) & \textbf{0.548} & \textbf{0.632} & \textbf{0.666} & \textbf{0.411} & \textbf{0.461} & \textbf{0.633} & \textbf{0.454} & \textbf{0.565} & \textbf{0.634} \\
\bottomrule
\end{tabular}
}

\vspace{0.5em}
\parbox{0.9\textwidth}{
\tiny
\textbf{Note:} The \textbf{best} and \underline{second-best} results for each metric are highlighted.
}
\end{table*}

\subsubsection{Error Attribution Analysis of SDM Pseudo-Labels}

To diagnose the primary sources of error in our SDM pipeline, we conducted a detailed error attribution analysis. Errors were categorized into two sources corresponding to the pipeline's core components: the segmentation model (SAM2 Error) and the vision-language model (CLIP Error). For SAM2, we report two metrics: False Segment (a generated mask with an IoU < 0.5 against any ground truth instance) and Mask Overlap. For CLIP, errors are quantified by False Negatives and False Positives.

The results are presented in Table \ref{tab:error_analysis}. As a quality benchmark, we first evaluated the pseudo-labels for the MegaStrawberry dataset against its manually annotated test set, achieving an mAP$_{50}$ of 0.603. This result validates the feasibility of using our pipeline to generate useful, large-scale training data, helping to address data scarcity in this domain. The analysis further reveals that the majority of errors across all datasets originate from the CLIP-based semantic matching stage, accounting for 90.40\%, 55.91\%,  65.25\%, and 65.43\% of the total errors for the MegaBlueberry, StrawDI\_Db1, MegaPeach, and MegaStrawberry datasets, respectively.

This finding suggests that the primary bottleneck for our SDM pipeline is the robustness of the vision-language alignment. Consequently, future work aimed at improving open-vocabulary segmentation performance would benefit most from enhancements in semantic matching capabilities.

\begin{table}[ht]
\centering
\caption{Quantitative analysis of pseudo-label quality and error attribution for the SDM pipeline.}
\label{tab:error_analysis}
\resizebox{\textwidth}{!}{
{
\begin{tabular}{@{}lrrcccccccc@{}}
\toprule
\multicolumn{3}{c}{\textbf{Dataset}} & \multicolumn{3}{c}{\textbf{Overall Precision}} & \multicolumn{2}{c}{\textbf{SAM2 Error}} & \multicolumn{2}{c}{\textbf{CLIP Error}} \\
\cmidrule(lr){1-3} \cmidrule(lr){4-6} \cmidrule(lr){7-8} \cmidrule(lr){9-10}
Name & Image & Instance & mAP$_{50:95}$ & mAP$_{50}$ & mAR & False Segment & Mask Overlap & False Negative & False Positive \\
\midrule
MegaStrawberry & 290 & 5,695 & 0.446 & 0.603 & 0.582 & 980 (34.57\%) & 0.794 & 740 (34.42\%) & 1,115 (39.33\%) \\
StrawDI\_Db1 & 2,800 & 16,234 & 0.548 & 0.632 & 0.666 & 2,703 (44.09\%) & 0.825 & 2,072 (34.42\%) & 1,245 (20.68\%) \\
MegaBlueberry & 1,778 & 12,898 & 0.411 & 0.461 & 0.633 & 902 (9.60\%) & 0.917 & 3,755 (34.42\%) & 4,738 (50.43\%) \\
MegaPeach & 1,680 & 6,349 & 0.454 & 0.565 & 0.634 & 1,003 (34.75\%) & 0.835 & 618 (34.42\%) & 1,265 (43.83\%) \\
\bottomrule
\end{tabular}
}}
\end{table}

\subsubsection{Cross-Category Generalization}
To evaluate the zero-shot cross-category generalization of SDM, we conducted qualitative experiments on categories beyond our primary datasets. We tested SDM on 16 common fruits ( Fig. \ref{fig1:SDM-D} and Fig. \ref{6fig:percep2}) and 8 challenging, uncommon crops (Fig. \ref{wired-fruits}). SDM exhibits a strong capacity for zero-shot instance segmentation across this diverse set of fruits. We further observed that SDM’s performance becomes less dependent on a specific class name and more sensitive to descriptive visual attributes (e.g., color and shape) provided in the prompt (see Section 4.5.2 and Appendix A.2 for more details). These findings suggest that SDM possesses a strong and flexible zero-shot generalization capability.

\begin{figure*}[ht]
    \centering
    \includegraphics[width=0.95\textwidth]{zero-shot1-eps-converted-to.pdf}
    \small
    \caption[short caption]{The results of SDM's zero-shot open-vocabulary segmentation on some common fruit images.}
    \label{6fig:percep2}
    \vspace{-0.3cm}
\end{figure*}

\subsection{Distilled Edge-deployable Models}

One of the key innovations of this work is the distillation of domain knowledge from large, computationally intense foundation models to smaller, edge-deployable student models. We compare the performance of students trained on pseudo-labels from our SDM against those trained on labels from three SOTA baselines, as well as a student trained on manual annotations which serves as an upper-bound reference. It is worth emphasizing that this distillation pipeline is designed to be model-agnostic, allowing practitioners to select an optimal student architecture that balances the accuracy and efficiency demands of their specific downstream task.

\subsubsection{Quantitative Performance Evaluation}
We evaluated the quality of SDM-generated pseudo-labels by using them to train various student models for object detection, semantic segmentation, and instance segmentation. For each task, student models were initialized with official pre-trained weights and trained using their default hyperparameters with no layers frozen. The comprehensive results are presented in Table \ref{3tab:distill_results}. For both object detection and semantic segmentation, the models distilled from SDM consistently outperform those distilled from other open-vocabulary baselines. Notably, the performance of our student models is highly competitive with the manually trained reference models, reaching up to 90.6\% of the mAP$_{50:95}$ score for object detection.

To provide a more rigorous validation for the instance segmentation task, each distillation experiment using YOLOv8s as the student was repeated five times with different random initialization. As detailed in Table \ref{3tab:distill_results}, the student model distilled via our SDM method demonstrates a clear and statistically significant superiority over three SOTA open-vocabulary baselines. This is confirmed by paired t-tests (p < $10^{-4}$), with the significance denoted by the † symbol. The performance gain is particularly pronounced on the MegaBlueberry dataset, where our model's mAP$_{50:95}$ is more than double that of the student distilled from Grounded SAM2 (0.678 vs. 0.337).

Overall, these statistically validated results confirm that our SDM-D pipeline generates superior pseudo-labels, enabling the training of robust student models that significantly outperform those distilled from other leading open-vocabulary methods.

\begin{table*}[!t]
\centering
\footnotesize
\caption{The performance of distilled models on three tasks.}
\label{3tab:distill_results}
\resizebox{\textwidth}{!}{%
\begin{tabular}{@{}ll ccc ccc ccc@{}}
\toprule
\multicolumn{11}{c}{\textbf{Results of the distilled models for object detection}} \\
\midrule
\textbf{Teacher model} & \textbf{Student model} & \multicolumn{3}{c}{\textbf{StrawDI\_Db1}} & \multicolumn{3}{c}{\textbf{Mega\_Blueberry}} & \multicolumn{3}{c}{\textbf{Mega\_Peach}} \\
\cmidrule(lr){3-5} \cmidrule(lr){6-8} \cmidrule(lr){9-11}
& & mAP$_{50:95}$ & mAP$_{50}$ & mAR & mAP$_{50:95}$ & mAP$_{50}$ & mAR & mAP$_{50:95}$ & mAP$_{50}$ & mAR \\
\midrule
Manual* & \multirow{5}{*}{\textbf{YOLOv8s}} & 0.826 & 0.937 & 0.846 & 0.781 & 0.880 & 0.844 & 0.781 & 0.921 & 0.843 \\
Grounded SAM & & 0.369 & 0.542& 0.620 & 0.395 & 0.441 & 0.544 & 0.542 & 0.716 & \underline{0.641} \\
Grounded SAM2 & & \underline{0.392} & \underline{0.598} & \underline{0.631} & 0.414 & 0.450 & 0.551 & \underline{0.558} & \underline{0.724} & \underline{0.648} \\
YOLO-World & & 0.352 & 0.469 & 0.618 & \underline{0.397} & \underline{0.534} & \underline{0.616} & 0.421 & 0.645 & 0.471 \\
SDM (Ours) & & \textbf{0.701} & \textbf{0.836} & \textbf{0.743} & \textbf{0.679} & \textbf{0.785} & \textbf{0.817} & \textbf{0.708} & \textbf{0.840} & \textbf{0.801} \\
\midrule
Manual* & \multirow{5}{*}{\textbf{Efficient-Det-D2}} & 0.738 & 0.879 & 0.778 & 0.731 & 0.865 & 0.846 & 0.668 & 0.822 & 0.779 \\
Grounded SAM & & 0.306 & 0.536 & 0.541 & 0.361 & 0.432 & 0.678 & 0.480 & 0.607 & 0.603 \\
Grounded SAM2 & & \underline{0.339} & \underline{0.551} & \underline{0.557} & \underline{0.380} & 0.451 & \underline{0.688} & \underline{0.492} & \underline{0.619} & \underline{0.611} \\
YOLO-World & & 0.272 & 0.453 & 0.554 & 0.357 & \underline{0.535} & 0.594 & 0.365 & 0.584 & 0.442 \\
SDM (Ours) & & \textbf{0.640} & \textbf{0.776} & \textbf{0.699} & \textbf{0.551} & \textbf{0.658} & \textbf{0.799} & \textbf{0.643} & \textbf{0.794} & \textbf{0.741} \\
\midrule[\heavyrulewidth]

\multicolumn{11}{c}{\textbf{Results of the distilled models for semantic segmentation}} \\
\midrule
\textbf{Teacher model} & \textbf{Student model} & \multicolumn{3}{c}{\textbf{StrawDI\_Db1}} & \multicolumn{3}{c}{\textbf{Mega\_Blueberry}} & \multicolumn{3}{c}{\textbf{Mega\_Peach}} \\
\cmidrule(lr){3-5} \cmidrule(lr){6-8} \cmidrule(lr){9-11}
& & Class Acc. & mIOU & FWIOU & Class Acc. & mIOU & FWIOU & Class Acc. & mIOU & FWIOU \\
\midrule
Manual* & \multirow{5}{*}{\textbf{DeepLab-v3+}} & 0.980 & 0.959 & 0.989 & 0.927 & 0.865 & 0.912 & 0.963 & 0.929 & 0.973 \\
Grounded SAM & & 0.945 & 0.835 & 0.952 & 0.603 & 0.533 & 0.711 & 0.866 & 0.838 & 0.940 \\
Grounded SAM2 & & \underline{0.948} & \underline{0.851} & \underline{0.963} & 0.618 & 0.551 & 0.720 & 0.871 & 0.846 & 0.945 \\
YOLO-World & & 0.875 & 0.786 & 0.938 & \underline{0.666} & \underline{0.583} & \underline{0.751} & \underline{0.942} & \underline{0.897} & \underline{0.960} \\
SDM (Ours) & & \textbf{0.966} & \textbf{0.946} & \textbf{0.986} & \textbf{0.830} & \textbf{0.757} & \textbf{0.848} & \textbf{0.949} & \textbf{0.898} & \textbf{0.961} \\
\midrule[\heavyrulewidth]

\multicolumn{11}{c}{\textbf{Results of the distilled models for instance segmentation}} \\
\midrule
\textbf{Teacher model} & \textbf{Student model} & \multicolumn{3}{c}{\textbf{StrawDI}} & \multicolumn{3}{c}{\textbf{Mega\_Blueberry}} & \multicolumn{3}{c}{\textbf{Mega\_Peach}} \\
\cmidrule(lr){3-5} \cmidrule(lr){6-8} \cmidrule(lr){9-11}
& & mAP$_{50:95}$ & mAP$_{50}$ & mAR & mAP$_{50:95}$ & mAP$_{50}$ & mAR & mAP$_{50:95}$ & mAP$_{50}$ & mAR \\
\midrule
Manual* & \multirow{5}{*}{\textbf{YOLOv8s}} & 0.773$\pm$0.013 & 0.929$\pm$0.009 & 0.788$\pm$0.003 & 0.760$\pm$0.010 & 0.895$\pm$0.010 & 0.833$\pm$0.009 & 0.891$\pm$0.008 & 0.911$\pm$0.006 & 0.848$\pm$0.006 \\
Grounded SAM & & 0.439$\pm$0.009 & 0.605$\pm$0.013 & 0.606$\pm$0.014 & 0.312$\pm$0.008 & 0.427$\pm$0.007 & 0.537$\pm$0.009 & 0.508$\pm$0.007 & 0.714$\pm$0.010 & 0.631$\pm$0.009 \\
Grounded SAM2 & & \underline{0.528}$\pm$0.005 & \underline{0.637}$\pm$0.007 & \underline{0.671}$\pm$0.007 & 0.337$\pm$0.010 & 0.428$\pm$0.011 & 0.563$\pm$0.003 & \underline{0.529}$\pm$0.011 & \underline{0.733}$\pm$0.007 & \underline{0.651}$\pm$0.007 \\
YOLO-World & & 0.231$\pm$0.016 & 0.401$\pm$0.009 & 0.444$\pm$0.010 & \underline{0.456}$\pm$0.006 & \underline{0.558}$\pm$0.008 & \underline{0.573}$\pm$0.008 & 0.519$\pm$0.019 & 0.732$\pm$0.006 & 0.641$\pm$0.007 \\
\textbf{SDM (Ours)} & & \textbf{0.682}$\pm$0.012$^{\dagger}$ & \textbf{0.838}$\pm$0.010$^{\dagger}$ & \textbf{0.691}$\pm$0.003$^{\dagger}$ & \textbf{0.678}$\pm$0.014$^{\dagger}$ & \textbf{0.780}$\pm$0.005$^{\dagger}$ & \textbf{0.720}$\pm$0.005$^{\dagger}$ & \textbf{0.607}$\pm$0.010$^{\dagger}$ & \textbf{0.804}$\pm$0.006$^{\dagger}$ & \textbf{0.749}$\pm$0.005$^{\dagger}$ \\
\bottomrule
\end{tabular}
}
\vspace{0.5em}
\parbox{0.9\textwidth}{
\tiny
\textbf{Note:} The \textbf{best} and \underline{second-best} results (excluding the manual baseline) are highlighted. The asterisk (*) denotes results from training with manual labels, which function as an upper-bound baseline. For the instance segmentation task, all results are expressed as mean $\pm$ standard deviation over five runs. The symbol $^{\dagger}$ indicates that our SDM model achieves a statistically significant improvement over all other teacher-models (Grounded SAM, Grounded SAM2, and YOLO-World) with $p < 10^{-4}$ in a paired t-test.
}
\end{table*}

\subsubsection{Foundation Model Versus Distilled Model}

\textbf{Inference Efficiency.} To quantify the inference efficiency of our proposed methods, we conducted a series of runtime analyses on both a high-end desktop workstation (NVIDIA RTX 4090 with 24 GB of VRAM) and a resource-constrained edge device (NVIDIA Jetson Orin NX with 16 GB of shared memory). For all benchmarks, we report Throughput (FPS), Inference Time (ms), and GPU Memory Usage (MiB). To ensure robust measurements, each experiment was repeated 10 times on 100 images randomly selected from StrawDI\_Db1 (batch size = 1), and the inference time is reported as mean ± standard deviation.

First, to provide a practical guide for implementation, we benchmarked the core SDM pipeline under various point\_grid and resolution settings on the desktop workstation (NVIDIA RTX 4090). As detailed in Table \ref{tab:sdm_inference_comparison}, the results allow practitioners to configure SDM for their specific needs.
Next, we conducted a comparative benchmark on the RTX 4090 to compare our full SDM pipeline, the distilled SDM-D student model (YOLOv8s), and three other foundation model-based baselines. As summarized in Table \ref{tab:sdm_inference_comparison}, the efficiency gains from distillation are dramatic: the student model reduces inference time by an exceptional 99.71\% compared to the original SDM. Finally, to validate its real-world deployability, we evaluated the lightweight SDM-D student model on the Jetson edge device. It achieved a throughput of 16.6 FPS (PyTorch) and 18.9 FPS (TensorRT-optimized), confirming its capability for real-time processing. This performance is a crucial enabler for deploying advanced perception on mobile platforms like agricultural robots, where on-board computation is essential for tasks such as automated harvesting and crop monitoring.

\begin{table*}[!t]
\centering
\footnotesize
\caption{\footnotesize Comparison of inference time and GPU memory usage for SDM across different settings.}
\label{tab:sdm_inference_comparison}
\resizebox{0.7\textwidth}{!}{%
\begin{tabular}{@{}lcccc@{}}
\toprule
\multicolumn{5}{c}{Comparison of metrics for SDM across different point\_grid (the resolution of used image is 1008$\times$756)} \\
\midrule
\textbf{Point\_grid} & \textbf{8} & \textbf{16} & \textbf{32} & \textbf{64} \\
\midrule
Throughput (FPS) & 1.264 & 0.463 & 0.212 & 0.057 \\
Inference Time$\pm$std (ms) & 790.88$\pm$0.91 & 2158.23$\pm$2.15 & 4723.33$\pm$3.04 & 17582.86$\pm$10.78 \\
GPU Memory Usage (MiB) & 6,082 & 6,852 & 6,940 & 7,788 \\
\midrule[\heavyrulewidth] 
\multicolumn{5}{c}{\textbf{Comparison of metrics for SDM across different resolutions (the point\_grid is set as 32)}} \\
\midrule
\textbf{Resolution} & \textbf{320$\times$320} & \textbf{640$\times$640} & \textbf{1008$\times$756} & \textbf{1024$\times$1024} \\
\midrule
Throughput (FPS) & 0.258 & 0.226 & 0.212 & 0.172 \\
Inference Time$\pm$std (ms) & 3,871.44$\pm$3.72 & 4,429.06$\pm$4.08 & 4,723.33$\pm$3.04 & 5,805.22$\pm$4.13 \\
GPU Memory Usage (MiB) & 6,246 & 6,616 & 6,940 & 7,642 \\
\hline
\end{tabular}
}
\end{table*}

\begin{table*}[!t]
\centering
\caption{\footnotesize Comparison of inference time and GPU memory usage for each method on the instance segmentation task.}
\label{tab:inference_comparison_gpus}
\resizebox{0.98\textwidth}{!}{%
\begin{tabular}{l|ccccc|cc}

\toprule
& \multicolumn{5}{c}{\textbf{NVIDIA RTX 4090 GPU}} & \multicolumn{2}{c}{\textbf{NVIDIA Jetson Orin NX Developer Kit}} \\
\cmidrule(lr){2-6} \cmidrule(lr){7-8}
\  & Grounded SAM & Grounded SAM2 & YOLO-World & SDM & SDM-D & SDM-D (PyTorch) & SDM-D (TensorRT) \\
\midrule
Throughput (FPS) & 0.329 & 0.755 & 17.99 & 0.212 & \textbf{55.096} & 16.625 & 18.854 \\
Inference Time$\pm$std (ms) & 3035.35$\pm$2.43 & 1325.15$\pm$3.55 & 55.58$\pm$1.40 & 4723.33$\pm$3.04 & \textbf{13.61$\pm$0.18} & 60.15$\pm$0.52 & 53.04$\pm$0.62 \\
GPU Memory Usage (MiB) & 7,574 & 4,329 & 1,988 & 6,940 & \textbf{754} & 881 & 874 \\
\bottomrule
\end{tabular}
}
\tiny
\textbf{Note:} The result in \textbf{bold} is the optimal one.
\end{table*}

\begin{figure}[ht]
    \centering
    \includegraphics[width=0.995\textwidth]{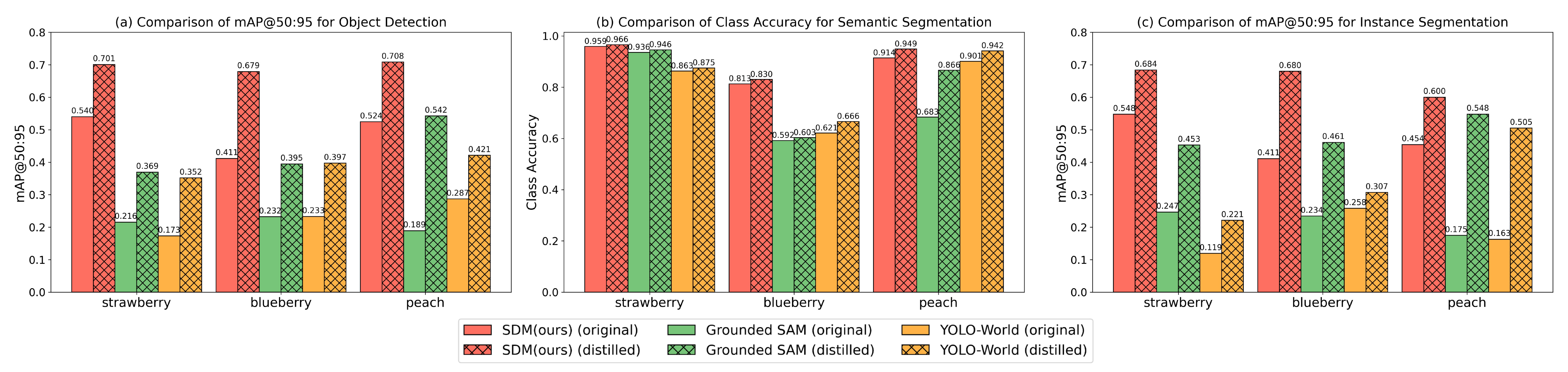}
    \small
    \caption[short caption]{Comparison of foundation models and distilled models. (a) Comparison of mAP@50:95 for object detection. (b) Comparison of Class Accuracy for semantic segmentation. (c) Comparison of mAP@50:95 for instance segmentation. For clarity, we only visualize the results of Grounded SAM, as Grounded SAM2 followed a similar trend.}
    \label{fig7:distill3}
    \vspace{-0.3cm}
\end{figure}

\textbf{Accuracy.} To evaluate the impact of distillation on performance, we compared the distilled student models to their foundation model teachers across three tasks, with results presented in Fig. \ref{fig7:distill3}. Surprisingly, we observe that the distillation process consistently improved perception accuracy. We term this phenomenon a “distillation improvement,” and notably, it occurred not only for our SDM but also for all tested baselines, including the Grounded SAM series and YOLO-World.
The magnitude of this improvement varied by task: it was most significant for object detection and instance segmentation (Fig. \ref{fig7:distill3}(a) and \ref{fig7:distill3}(c)) nd more modest for semantic segmentation (Fig. \ref{fig7:distill3}(b)). We hypothesize two potential reasons for this effect: (1) the distillation process implicitly focuses the model on the target domain's data distribution, and (2) the student model training process may average out noise from the imperfect pseudo-labels, leading to a more robust result. This counter-intuitive improvement through distillation has also been observed in prior work \cite{51-xie2020selftrainingnoisystudentimproves}.

\subsection{Fine-tuning with Few Manual Labels}

While our SDM-D pipeline produces effective zero-shot models, a fine-tuning step with a few manual labels can bridge and align the model with specific annotation criteria (e.g., whether a "strawberry" label should include the calyx). We evaluated both object detection and instance segmentation on the StrawDI\_Db1 dataset. Specifically, the student models were fine-tuned with 1, 50, and 100 labeled images randomly drawn from the training set, with results averaged over 10 runs to reduce sampling bias. Fine-tuning was conducted on YOLOv8s (distilled from SDM-D) without freezing any layers. In the 1-shot setting, the batch size was 1 and training lasted for 150 epochs, while in the 50-shot and 100-shot settings, the batch size was 16 and training lasted for 250 epochs. For comparison, models trained from scratch were constructed using 1-2,900 labeled images, increasing in steps of 5 for the first 50 images and 50 thereafter, with samples randomly selected from the full training set. Furthermore, to benchmark our method against traditional annotation-light techniques, we implemented a semi-supervised learning (SSL) baseline on the instance segmentation task. In this SSL setup, a YOLOv8s model is first trained using manual label sets of varying sizes, starting at 5 and 10, followed by increments of 10 up to 50, and then increments of 50 up to 300. This model then generates pseudo-labels for the remaining unlabeled images, which are combined with the initial manual set to retrain the model. This allows us to directly compare the data efficiency of fine-tuning our pre-trained model versus using a popular SSL strategy, with the results for each data point averaged over five runs with different random initialization.

The results, illustrated in Fig. \ref{fig8-fitunning}, highlight the significant data efficiency of our approach. Our initial zero-shot distilled model already performs comparably to a model trained from scratch with approximately 200 manual labels. Remarkably, fine-tuning with just one labeled image boosts the performance to over 91\% of the fully supervised model. To match the performance of our 1-shot and 50-shot fine-tuned models, a from-scratch approach requires roughly 250 and 1,050 labeled images, respectively. To demonstrate the general applicability of this benefit, we replicated these findings with other SOTA student architectures prevalent in agricultural vision \cite{28-article} (Fig. \ref{fig8-fitunning}(c-d)). These experiments consistently show that models pre-trained via SDM-D require only a single manually labeled image to achieve over 90\% of the performance of their fully supervised counterparts trained on thousands of labels.

Furthermore, our fine-tuning strategy demonstrates clear advantages over a traditional Semi-Supervised Learning (SSL) baseline, particularly in low-data regimes, as illustrated by the "Human SSL" curve in Fig. \ref{fig8-fitunning}(b). Although our zero-shot model only surpasses the SSL baseline trained with 30 labeled images (0.668 vs. 0.650), after fine-tuning with a single labeled image it can already match the performance of the SSL model trained with 100 labels (0.708 vs. 0.709). This indicates that the powerful zero-shot initialization from SDM-D provides a much more effective starting point for few-shot learning than iterative pseudo-labeling from a weakly trained model. Ultimately, this offers a scalable and cost-effective solution for adapting perception systems to new crops or environments, addressing the critical bottleneck of labor-intensive manual annotation.

\begin{figure*}
\centering
    \includegraphics[width=0.75\textwidth]{Figure_8-new-eps-converted-to.pdf}
    \small
    \caption[short caption]{Comparison of the student model’s few-shot learning results with full training results on manual data. (a) Object detection with YOLOv8s as the student model. (b) Instance Segmentation with YOLOv8s as the student model. (c) Object detection with YOLOv8s, YOLOv11, and Mango-YOLO as student models. (d) Instance Segmentation with YOLOv8s, YOLOv11, and Mask R-CNN as student models. All error bars indicate 95\% confidence intervals (CI).}
    \label{fig8-fitunning}
    \vspace{-0.3cm}
\end{figure*}

\subsection{Ablation study}
\subsubsection{Hyperparameter Sensitivity Analysis}

We conducted a sensitivity and ablation study to validate our hyperparameters and the Mask-NMS module. Experiments used 100 random StrawDI images, varying one parameter (point-grid size, Mask-NMS threshold, or min area) while holding others at their defaults (32, 0.9, and 50, respectively). Results are shown in Fig. \ref{ablation-hyp}.

\begin{figure*}
    \centering
    \includegraphics[width=0.8\textwidth]{ablation1-esp-converted-to.pdf}
    \small
    \caption[short caption]{Hyperparameter Sensitivity of SDM on instance segmentation task. (a) Effect of different point-grid settings with and without Mask-NMS. (b) AP performance under different Mask-NMS thresholds. (c) AP performance under different minimum area filters.}
    \label{ablation-hyp}
    \vspace{-0.3cm}
\end{figure*}
Specifically, Fig. \ref{ablation-hyp}(a) serves a dual purpose: it presents the ablation of Mask-NMS by showing performance with and without the module across four different point-grid densities (8, 16, 32, and 64), which simultaneously reveals the model's sensitivity to grid size. As shown in Fig. \ref{ablation-hyp}(a), applying Mask NMS consistently improved AP across different grid settings, indicating its effectiveness in reducing duplicate predictions. The benefits of this module are further substantiated on the MegaBlueberry and MegaPeach datasets (see Appendix A.2 for details). We also analyzed the sensitivity to the Mask NMS threshold (Fig. \ref{ablation-hyp}(b)) and minimum area filter (Fig. \ref{ablation-hyp}(c)) for more references.

\subsubsection{Prompts Design}

According to the experiment, we find that the design of prompts has a substantial impact on model performance. A large portion of labeling errors originates from the region–text matching step, and carefully designed prompts can effectively mitigate such errors. As shown in Table \ref{tab:prompt_comparison}, using a richer set of prompts that comprehensively cover the main scene elements (e.g., fruits, leaves, stems, and flowers) improves segmentation accuracy by reducing confusion between visually similar regions. Moreover, attribute-enriched prompts provide stronger discriminative cues than plain labels.

\begin{table*}[!t]
\centering
\caption{\footnotesize The results of instance segmentation on 100 StrawDI images with different prompts.}
\label{tab:prompt_comparison}
\resizebox{0.8\textwidth}{!}{%
\begin{tabular}{@{}l >{\raggedright\arraybackslash}p{8cm} ccc@{}}
\toprule
 & \textbf{prompts} & \textbf{mAP$_{50:95}$} & \textbf{mAP$_{50}$} & \textbf{mAR} \\
\midrule
6 complex labels & a red strawberry; a pale green strawberry with numerous points; a green veined leaf; a long and thin stem; a white flower; soil or background or something else. & 0.523 & 0.607 & 0.653 \\
\addlinespace 
5 complex labels & a red or green strawberry; a green veined leaf; a long and thin stem; a white flower; soil or background or something else. & 0.507 & 0.579 & 0.558 \\
\addlinespace
5 simple labels & strawberry; leaf; flower; soil or background. & 0.473 & 0.536 & 0.525 \\
\addlinespace
2 labels & strawberry; leaf or background not a strawberry. & 0.374 & 0.429 & 0.462 \\
\bottomrule
\end{tabular}
}
\end{table*}

We summarize an effective prompt template: a/an \{color\} \{shape\} \{object\} with \{feature\}. Among them, the color description is the most crucial. As illustrated in Fig. \ref{fig:prompt-effect}(a) and Fig. \ref{fig:prompt-effect}(b),
“fix-it” prompts ("New texts" highlighted in pink boxes) effectively suppress common false positives in complex, densely cluttered agriculture scenes. These refined prompts explicitly describe the background color (e.g., “pale flesh-coloured background” for the peach scene or “white pale blue background” for the apple scene), which helps the model better distinguish target fruits from visually similarly colored, non-target regions. While Fig. \ref{fig:prompt-effect}(c) highlights typical errors (orange arrows) that could sometimes be corrected by introducing specific background descriptions (e.g., “black background”). Nevertheless, to preserve generality across datasets, we avoided overly scene-specific prompts. Finally, while increasing the number and complexity of prompts generally leads to higher accuracy, excessive prompt diversity may harm model generalization on large-scale datasets and impose considerable annotation costs. We therefore recommend balancing prompt expressiveness with dataset characteristics when applying SDM.

\begin{figure*}[ht]
    \centering
    \includegraphics[width=0.75\textwidth]{Figure_10-eps-converted-to.pdf}
    \small
    \caption[short caption]{Influence of prompts on label assignment. (a) and (b) illustrate representative fix-it prompts that improve label assignment accuracy in dense agricultural scenes. (c) Examples of matching between segmentation masks and prompts of strawberry. The numbers in bold represent the highest similarity. The numbers with an underline is the second-highest similarity. The figures on the left are the results of old prompts, and on the right are the results of new prompts.}
    \label{fig:prompt-effect}
    \vspace{-0.3cm}
\end{figure*}

\section{Conclusion}

This paper presents an innovative framework, SDM-D. The primary contribution of this work is the establishment of a comprehensive framework that leverages the knowledge within pre-trained foundation models for fruit perception and distills this knowledge into edge-deployable models that excel in both speed and accuracy. A key component is our Mask Crop and Mask NMS strategies, which have been proven highly effective for robust perception in dense, complex scenes. Extensive experiments and rigorous statistical analysis demonstrate that our method consistently surpasses SOTA OVD methods across various fruit scenes. The distilled student model demonstrates high efficiency, running at 18.9 FPS on an edge device. In terms of accuracy, it provides a strong zero-shot performance at 86.6\% of the fully supervised reference model, which is boosted to 91.8\% with only a single labeled image for fine-tuning. 

Our SSL experiments confirm that our framework yields notable savings in manual instance-level annotation by offering a stronger supervisory signal in low-data regimes, thereby reducing the cost and time required to develop high-performance fruit perception models. This advancement accelerates the development and deployment of agricultural robots, enhancing efficiency and scalability in tasks such as fruit monitoring and harvesting. Furthermore, we believe this approach holds potential applications beyond agriculture, extending to fields such as healthcare, autonomous driving, and robotics, where open vocabulary segmentation is required. We also present a high-quality dataset, MegaFruits, which includes over 25,000 annotated images of strawberries, peaches, and blueberries, making it the largest open fruit segmentation dataset. We hope this resource can advance research and applications in agricultural perception. 

There are still some limitations in SDM-D. First, the performance of the distilled student models, while competent, does not yet match that of models trained on extensive, fully-supervised datasets. And the distilled student models with high inference speed, show limited adaptability compared to their foundation model teachers, requiring redistillation as environmental conditions or task requirements change. Second, our pipeline's effectiveness is largely dependent on the capabilities of the pre-trained foundation models (SAM2 and OpenCLIP). Errors or biases from these models, especially in scenarios unseen during their training, can propagate into the pseudo-labels and subsequently into the student model. Similarly, SDM is designed for generality and does not require any training, which contributes to its convenience but may impact accuracy when handling domains requiring specialized annotations (see Appendix A.1). Moreover, while we eliminate per-instance annotation, the system is not fully autonomous, as performance remains sensitive to the expert-driven design of text prompts. Regarding the scope of our evaluation, our comparisons focused primarily on other OVD methods, and a broader benchmark against other annotation-light paradigms (e.g., active learning) was not conducted. Additionally, while we presented qualitative generalization results on less common crops, a lack of suitable public datasets prevented a rigorous quantitative analysis.

Looking ahead, these limitations highlight several exciting research avenues. Key challenges include enhancing the pipeline's robustness, possibly through efficient fine-tuning of the foundation models, and improving the adaptability of student models to new environments without requiring full re-distillation. Investigating the mechanisms behind the observed "distillation improvement" phenomenon is a compelling avenue for future research. Finally, developing methods for automatic prompt generation and explore more sophisticated cross-modal alignment techniques that go beyond cosine similarity can enhance the model's robustness and performance in complex, real-world scenes.

\section*{Declaration of Competing Interest}
The authors declare that they have no known competing financial interests or personal relationships that could have appeared to influence the work reported in this paper.

\section*{Acknowledgements}
This study was financially supported by the National Natural Science Foundation of China (grant number U20A2019).

\appendix

\renewcommand{\thefigure}{A.\arabic{figure}}  
\renewcommand{\thetable}{A.\arabic{table}}
\setcounter{figure}{0}
\setcounter{table}{0}

\section*{Appendix A.1 Performance on Datasets with Specialized Annotation Protocols}

We evaluated OVD instance segmentation on the MinneApple dataset, which uses a specialized protocol that annotates only on-tree apples. This task proved challenging for all methods (Table \ref{tab:minneapple_results}). While our SDM achieved the highest performance, its mAP$_{50:95}$ was only 0.166. This poor result across all models highlights the difficulty of applying general OVD approaches directly to domains with highly specialized labeling schemes.

\begin{table*}[!t]
\centering
\caption{\footnotesize Instance segmentation results of OVD methods on the MinneApple dataset.}
\label{tab:minneapple_results}
\resizebox{0.6\textwidth}{!}{%
\begin{tabular}{lcccc} 
\toprule
 & \textbf{Grounded SAM} & \textbf{Grounded SAM2} & \textbf{YOLO-World} & \textbf{SDM} \\
\midrule
mAP$_{50:95}$ & 0.116 & \underline{0.125} & 0.118 & \textbf{0.166} \\
\addlinespace
mAP$_{50}$ & 0.274 & 0.280 & \underline{0.329} & \textbf{0.344} \\
\addlinespace
mAR$_{50:95}$ & 0.271 & 0.281 & \underline{0.293} & \textbf{0.298} \\
\bottomrule
\end{tabular}
}
\end{table*}

\section*{Appendix A.2 SDM on "unseen" crop}
We test SDM on 8 less common plants likely underrepresented in the foundation models’ training data. As shown in the Fig. \ref{wired-fruits}, SDM works well under challenging conditions, such as targets with varied morphology or colors that closely match the background.

\begin{figure*}[ht]
    \centering
    \includegraphics[width=0.75\textwidth]{wired-fruits-eps-converted-to.pdf}
    \small
    \caption[short caption]{The zero-shot open-vocabulary segmentation results of SDM on less common fruit species with diverse shapes and appearances. (a) Papaya: a tropical fruit whose skin color is often indistinguishable from the background. (b) Baobab: large woody fruit with irregular contours. (c) Cassava: a root crop whose color closely matches the surrounding soil. (d) Cocoa: pods with varying ripeness and natural camouflage. (e) Cactus: Fruits with diverse colors and distinctive morphology.}
    \label{wired-fruits}
    \vspace{-0.3cm}
\end{figure*}

\section*{Appendix A.3 Ablation of Mask Crop and Mask NMS}
We evaluated the impact of the Mask Crop and Mask NMS components within the SDM pipeline across three datasets and three tasks. The results, summarized in Table \ref{tab:ablation_mask_crop_nms}, demonstrate the effectiveness of both components in dense scenes.

\begin{table}[!t]
\centering
\footnotesize
\caption{The ablation study results (mAP$_{50:95}$) of Mask Crop and Mask NMS across object detection, semantic segmentation and instance segmentation tasks.}
\label{tab:ablation_mask_crop_nms}
\resizebox{\textwidth}{!}{
\begin{tabular}{ll ccc ccc ccc}
\toprule
\multicolumn{2}{c}{\textbf{Component}} & \multicolumn{3}{c}{\textbf{StrawDI\_Db1}} & \multicolumn{3}{c}{\textbf{MegaBlueberry}} & \multicolumn{3}{c}{\textbf{MegaPeach}} \\
\cmidrule(r){1-2} \cmidrule(lr){3-5} \cmidrule(lr){6-8} \cmidrule(l){9-11}
\textbf{Mask Crop} & \textbf{Mask NMS} & \textbf{Object Det.} & \textbf{Semantic Seg.} & \textbf{Instance Seg.} & \textbf{Object Det.} & \textbf{Semantic Seg.} & \textbf{Instance Seg.} & \textbf{Object Det.} & \textbf{Semantic Seg.} & \textbf{Instance Seg.} \\
\midrule
No  & Yes & 0.409 & 0.897 & 0.412 & 0.372 & 0.805 & 0.370 & 0.492 & 0.903 & 0.425 \\
Yes & No  & 0.487 & 0.923 & 0.493 & 0.360 & 0.789 & 0.362 & 0.847 & 0.892 & 0.404 \\
Yes & Yes & \textbf{0.540} & \textbf{0.959} & \textbf{0.548} & 0.411 & 0.813 & 0.411 & 0.524 & 0.914 & 0.454 \\
\bottomrule
\end{tabular}}
\end{table}

\section*{Appendix A.4 Statistical Analysis}
To supplement the results in Table 5.3, we performed a detailed statistical analysis on the paired (N=5 seeds) instance segmentation data. This rigorous procedure included: (1) calculation of the effect size (paired Cohen's d, $d_z$
); (2) calculation of the 95\% Confidence Intervals (CI) of the mean difference; (3) an omnibus test using one-way repeated measures ANOVA (rmANOVA), which confirmed significant overall differences among methods on all datasets (p <0.0001); and (4) post-hoc paired t-tests comparing SDM to the four baselines, with the Holm-Bonferroni correction applied to address multiple comparisons (yielding the adjusted p-value, $p_{adj}$). The complete results in Table A.2 confirm that the superior performance of our SDM framework over the non-manual baselines (Grounded SAM, Grounded SAM2, and YOLO-World) is highly statistically significant (all $p_{adj}$ < 0.01).

\begin{table*}[!t]
\centering
\footnotesize
\caption{Detailed statistical comparison of SDM (ours) vs. baselines on the instance segmentation task.}
\label{tab:appendix_stats}
\resizebox{\textwidth}{!}{%
\begin{tabular}{l ccc ccc ccc}
\toprule
\multirow{2}{*}{\textbf{Teacher model}} & \multicolumn{3}{c}{\textbf{StrawDI\_Db1}} & \multicolumn{3}{c}{\textbf{Mega\_Blueberry}} & \multicolumn{3}{c}{\textbf{Mega\_Peach}} \\
\cmidrule(lr){2-4} \cmidrule(lr){5-7} \cmidrule(lr){8-10}
 & Cohen's d ($d_z$) & 95\% CI [Diff] & p-value (adj) & Cohen's d ($d_z$) & 95\% CI [Diff] & p-value (adj) & Cohen's d ($d_z$) & 95\% CI [Diff] & p-value (adj) \\
\midrule
Manual* & -5.784 & [-0.1105, -0.0715] & 2.06103e-04 & -4.908 & [-0.1026, -0.0612] & 3.91832e-04 & -20.095 & [-0.3010, -0.2660] & 5.86833e-06 \\
Grounded SAM & 23.658 & [0.2302, 0.2558] & 2.29286e-06 & 24.984 & [0.3481, 0.3845] & 1.84405e-06 & 6.794 & [0.0811, 0.1174] & 3.28460e-04 \\
Grounded SAM2 & 15.227 & [0.1403, 0.1653] & 8.87796e-06 & 39.306 & [0.3308, 0.3524] & 4.01851e-07 & 5.913 & [0.0621, 0.0951] & 3.78156e-04 \\
YOLO-World & 40.095 & [0.4366, 0.4646] & 3.71153e-07 & 15.953 & [0.2048, 0.2393] & 7.37196e-06 & 3.684 & [0.0586, 0.1182] & 1.18402e-03 \\
\bottomrule
\end{tabular}
}
\vspace{0.5em}
\parbox{0.95\textwidth}{
\tiny

\textbf{Note:} Statistical analysis compares SDM (Ours) against each baseline (N=5 paired runs). 95\% CI [Diff] denotes the 95\% confidence interval of the mean difference. p-value (adj) reports the p-value after Holm-Bonferroni correction for multiple comparisons.}
\end{table*}

\section*{Appendix A.5 Experimental Configuration}
The detailed configuration settings used in our experiments are summarized in Table \ref{tab:appendix_prompts_params}

\begin{table*}[!t]
\centering
\footnotesize
\caption{Prompt templates and key parameters used for pseudo-label generation.}
\label{tab:appendix_prompts_params}
\resizebox{1\textwidth}{!}{%
\begin{tabular}{ccl}
\toprule
\textbf{Dataset} & \textbf{Prompts (Class: ``description'')} & \textbf{Parameters} \\
\midrule
\textbf{MegaStrawberry} & 
    \parbox{10cm}{\textbf{ripe:} a red strawberry; \textbf{unripe:} a white green strawberry with seeds; \textbf{leaf:} a green strawberry leaf; \textbf{stem:} a strawberry stem with green hairs; \textbf{others:} a white gray cement thing or background}
    & 
    \parbox{2.2 cm}{\raggedright Seeds: 0 (default), 329, 1107, 5804, 8899 } \\
\cmidrule(r){1-2}
\textbf{StrawDI\_Db1} & 
    \parbox{10cm}{ \textbf{strawberry(ripe):} a red strawberry; \textbf{strawberry(unripe):} a pale green strawberry with numerous points; \textbf{leaf:} a green veined strawberry leaf; \textbf{stem:} a long and thin stem; \textbf{flower:} a white flower; \textbf{others:} soil or background or something else}
    & 
    \parbox{2.2 cm}{\raggedright Point Grid: 32 × 32} \\  
\cmidrule(r){1-2}
\textbf{MegaBlueberry} & 
    \parbox{10cm}{\textbf{ripe:} a bluish violet blueberry with scratch; \textbf{unripe:} a short circle olive green or a pink blueberry with scratch; \textbf{leaf:} an oblong green leaf; \textbf{stem:} a slender thin twig pink or green; \textbf{others:} brown red background or others}
    & 
    \parbox{2.2 cm}{\raggedright MaskNMS: 0.9} \\  
\cmidrule(r){1-2}
\textbf{MegaPeach} & 
    \parbox{10cm}{\textbf{peach:} a yellow red peach; \textbf{leaf:} a green veined leaf; \textbf{others:} pale flesh-coloured background or stem or others}
    & 
    \parbox{2.2 cm}{\raggedright Mini-Area: 50} \\  
\bottomrule
\end{tabular}
}
\parbox{0.95\textwidth}{
\tiny

\textbf{Note:} When training student models, the labels "ripe"and "unripe" in StarwDI are unified into a class of "strawberry". (2) The “others” category in each dataset refer to background region not covered by the defined labels.}
\end{table*}

\section*{Appendix B. SDM-D Demonstration Link}

A public demonstration of the proposed \textbf{SDM-D} framework is available at: \noindent\url{https://www.kaggle.com/datasets/mmwang0/the-demos-of-sdm-d}




\bibliographystyle{elsarticle-num}
\bibliography{arxiv}

\end{document}